%% file: neurips_2026.tex
\documentclass{article}

\usepackage[preprint]{neurips_2026}

\usepackage[utf8]{inputenc} % allow utf-8 input
\usepackage[T1]{fontenc}    % use 8-bit T1 fonts
\usepackage{hyperref}       % hyperlinks
\usepackage{url}            % simple URL typesetting
\usepackage{booktabs}       % professional-quality tables
\usepackage{amsfonts}       % blackboard math symbols
\usepackage{nicefrac}       % compact symbols for 1/2, etc.
\usepackage{microtype}      % microtypography
\usepackage{xcolor}         % colors
\usepackage{graphicx}
\usepackage{multirow}
\usepackage{booktabs} % 用于 \toprule, \midrule 等精美横线
\usepackage{multirow} % 用于单元格跨行合并
\usepackage{xcolor}   % 用于 \textcolor{gray} 设置灰色字体
\usepackage[table]{xcolor} % 必须带 [table] 参数，否则不认识 \rowcolor
\usepackage{amsmath}
\usepackage{amssymb}
\usepackage{bm}
\usepackage{subcaption}
\usepackage{wrapfig}
\usepackage{booktabs}
\usepackage{tabularx}
\usepackage{makecell}
\usepackage{array}
\usepackage{tikz}
\newcommand{\tagnum}[1]{%
  \tikz[baseline=(char.base)]{
    \node[fill=gray!20, rounded corners=1.5pt, inner sep=2pt, font=\sffamily\bfseries\scriptsize] (char) {#1};
  }%
}
\newcommand{\mymethod}{FlexDraft}
\newcommand{\attntuning}{Attn Tuning}
\newcommand{\bonusbias}{Bonus-guided Calibration}
\newcommand{\flexdecoding}{Flex Decoding}

\title{FlexDraft: Flexible Speculative Decoding via Attention Tuning and Bonus-Guided Calibration}

\author{
    \textbf{Yaojie Zhang$^{1,3}$\thanks{Equal contribution. $^\dag$Corresponding author.} 
    \qquad Jianuo Huang$^{1,2*}$ \qquad Junlong Ke$^{4}$ \qquad Yuhang Han$^{1,5}$} \vspace{4pt}\\
    \textbf{Yongji Long$^{2}$ \qquad Tianchen Zhao$^{4}$ \qquad Biqing Qi$^{6}$ \qquad Linfeng Zhang$^{1\dag}$}\vspace{5pt} \\
    $^1$EPIC Lab, SJTU \qquad
    $^2$UESTC \qquad
    $^3$School of Software Engineering, HUST \\
    $^4$Tsinghua University \qquad
    $^5$HKUST(GZ) \qquad
    $^6$Shanghai AI Laboratory \vspace{2pt} \\
    \texttt{\small \{yaojiezhang288,jianuohuang82\}@gmail.com}
}

\begin{document}

\maketitle

\input{sec/1abstract}

\section{Introduction} \label{sec:intro}

\input{sec/2introduction}

\section{Related Work} \label{sec:related}
\input{sec/3related_work}

\section{Preliminaries: Speculative Decoding} \label{sec:preliminaries}
\input{sec/preliminaries}

\section{\mymethod{}} \label{sec:method}
\input{sec/4method}

\section{Experiments} \label{sec:exp}
\input{sec/5experiment}

\section{Discussion} \label{sec:discussion}
\input{sec/6discussion}

\section{Conclusion} \label{sec:conclusion}
\input{sec/7conclusion}

\bibliographystyle{plain}
\bibliography{references}

\newpage
\appendix
\input{sec/8appendix}

\end{document}

%% file: sec/1abstract.tex
\begin{abstract}
Speculative decoding accelerates memory-bound LLM inference without quality degradation by using a fast drafter to propose multiple candidate tokens and the target model to verify them in parallel.
However, conventional sequential speculative decoding suffers from mutual waiting between drafting and verification, and repeated exchange of intermediate states further increases memory access overhead.
Parallel speculative decoding addresses this limitation by performing drafting and verification within a single target forward pass, allowing future drafts to be prepared while current candidates are being verified.
Although effective at small batch sizes, existing parallel speculative decoding methods either require costly continual pretraining with quality degradation or suffer from low acceptance rates.
More importantly, this paradigm inherently suffers from uncertainty in both the bonus token and the accepted length, leading to draft verification mismatch and causing throughput gains to collapse at large batch sizes.
To address these limitations, we introduce \mymethod{}, a lossless speculative decoding framework that flexibly adapts to varying batch sizes through three key designs.
(1) Attention Tuning enables block diffusion drafting by tuning only the attention projectors of the final few layers on mask tokens, while keeping the autoregressive path frozen to exactly preserve the target distribution and produce high quality drafts with minimal trainable parameters.
(2) \bonusbias{} uses a lightweight MLP conditioned on the resolved bonus token to calibrate draft logits, mitigating the draft verification mismatch caused by bonus token uncertainty.
(3) \flexdecoding{} dynamically switches between parallel draft and verify at small batch sizes and sequential draft then verify at large batch sizes, and further adjusts verification length based on draft confidence to eliminate redundant computation.
Experiments on Qwen3-8B show that \mymethod{} achieves an average $4.59\times$ speedup over autoregressive decoding without quality loss.
\emph{Our code will be released on Github.}
\end{abstract}

%% file: sec/2introduction.tex
Autoregressive large language models (LLMs) have demonstrated remarkable capabilities across diverse natural language understanding and generation tasks \citep{team2024gemini, deepseekai2024deepseekv3technicalreport, singh2025openai}, but their token by token decoding paradigm makes inference highly memory bound and fundamentally limits generation speed.
Speculative decoding addresses this bottleneck as a lossless acceleration paradigm by using a lightweight drafter to propose multiple candidate tokens and the target model to verify them in a single forward pass, thereby amortizing expensive target model computation over multiple accepted tokens \citep{chen2023accelerating, leviathan2023fast}.
Prior methods improve speculative decoding by enhancing draft quality \citep{zhou2024distillspec, li2025eagle3}, reducing draft generation cost \citep{an2026pard, chen2026dflash}, or organizing candidates more effectively for verification \citep{miao2024specinfer, li2024eagle2}, with representative examples such as EAGLE for feature-level draft prediction \citep{li2024eagle, li2025eagle3} and DFlash for target conditioned block diffusion drafting \citep{chen2026dflash}.
% \vspace{-4mm}

Despite these advances, most conventional speculative decoding methods still follow a sequential draft then verify schedule, where the next draft cannot begin until verification completes. 
This serialized execution induces mutual waiting between the drafter and the target model, and repeated exchange of intermediate states further increases memory access overhead, limiting end-to-end speedup \citep{liu2025pearl, shen2026specbranch}.

\begin{figure*}[t]
  \centering
  % \vspace{-2mm}
  \includegraphics[width=1.0\linewidth]{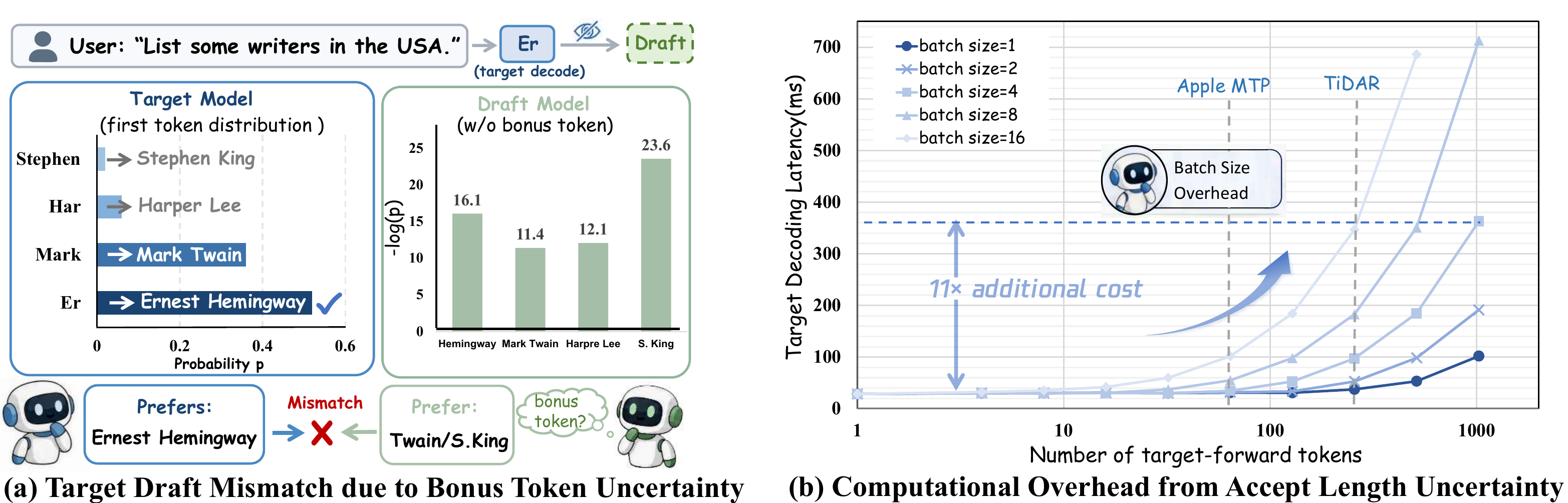}
  % \vspace{-1mm}
  \caption{\textbf{Limitations of the parallel speculative decoding paradigm.}
  (a) Given the prefix, the target model predicts the next token and prefers ``Er'', the first token of ``Ernest Hemingway'', which constrains the subsequent generation toward Hemingway. Without access to the bonus token, the draft model tends to favor alternative continuations.
  (b) Accept length uncertainty forces parallel speculative decoding to consider all potentially accepted prefixes of the last draft, requiring mask embeddings to traverse all target layers and causing the target forward token count to grow as $O(N^2)$. The right figure analyzes the resulting overhead for different numbers of target forward tokens, which is moderate at small batch sizes but grows rapidly at larger batch sizes.}
  \vspace{-6mm}
  \label{fig:bonus_uncertain}
\end{figure*}

To reduce this mutual waiting, recent methods prepare future candidates before verification is fully resolved, enabling parallel speculative decoding that overlaps drafting with verification within a single target forward pass \citep{lin2025bita, liu2025tidar, samragh2025your}.
However, existing parallel speculative decoding methods either require continual pretraining with substantial training and parameter overhead, potentially leading to quality degradation, or suffer from low acceptance rates.
More importantly, this paradigm introduces two fundamental limitations.
Since future drafts are generated before the verification outcome is known, the drafter faces uncertainty in both the bonus token and the accepted length.
These two uncertainties affect parallel future drafting in different ways.
Bonus token uncertainty weakens draft verification alignment, while acceptance length uncertainty forces the drafter to generate candidate branches for multiple possible accepted prefixes, introducing substantial redundant drafting overhead.
We detail these two limitations below.

\textbf{\tagnum{1}~Bonus token uncertainty.}
% \textbf{Bonus token uncertainty.}
In speculative decoding, when the first rejection occurs during verification, the target model samples a corrected token at the rejected position, which is commonly referred to as the \textit{bonus token}.
However, in parallel speculative decoding, this bonus token is not available when the draft tokens are generated \citep{lin2025bita, liu2025tidar}.
As a result, the draft model must predict future tokens without knowing which corrected token will be produced by the target model, leading to a mismatch between drafting and verification.
Figure~\ref{fig:bonus_uncertain}(a) illustrates this issue.
Given the query ``List some writers in the USA.'', the target model may produce \texttt{Er} as the bonus token, corresponding to the beginning of \texttt{Ernest Hemingway}.
Conditioned on this token, the subsequent verified trajectory should remain coherent with \texttt{Ernest Hemingway}.
In contrast, because the draft model cannot observe the bonus token during drafting, it may assign high probability to continuations associated with other writers, such as \texttt{Mark Twain} or \texttt{S. King}.
Although these drafts may be locally plausible under the original prefix, they become inconsistent once the verifier selects \texttt{Er} as the bonus token.
This mismatch reduces the probability that the drafted tokens can be accepted, thereby limiting the speedup of parallel speculative decoding.

\textbf{\tagnum{2}~Acceptance length uncertainty.}
In parallel draft and verify methods, future candidates are generated before the actual acceptance length is known \citep{liu2025tidar, samragh2025your, lin2025bita}.
Since the verifier determines the accepted length only after verification, the drafter cannot know which future branch will remain valid and must prepare candidates for multiple possible accepted prefixes, introducing substantial redundant computation.
In existing parallel speculative decoding methods, the additional mask tokens used for future drafting are injected from the bottom layers and must pass through all target layers together with the verification tokens.
Consequently, as the draft block length increases, covering all possible accepted prefixes requires a rapidly growing number of mask tokens, leading to an $O(N^2)$ increase in the effective number of target forward tokens.
This creates an inherent trade-off.
Methods with low acceptance rates usually adopt short draft blocks, because larger blocks bring limited accepted token gains but much higher mask token overhead.
Methods such as TiDAR can obtain longer accepted continuations through continual pretraining, but their larger blocks further amplify the redundant target forward computation caused by acceptance length uncertainty.
As shown in Figure~\ref{fig:bonus_uncertain}(b), the target side overhead exceeds $11\times$ at batch size 16.
Consequently, the latency savings from overlapping drafting and verification are ultimately offset or even outweighed by the extra target forward computation, causing speedup to collapse at large batch sizes.

To address these challenges, we introduce \mymethod{}, a speculative decoding framework that flexibly adapts to varying batch sizes without quality degradation, built on three key contributions. 
\textbf{(1) Attention Tuning (\attntuning{})} tunes only the attention projectors of the final few layers for mask token prediction, while keeping the autoregressive verification path frozen. 
It adds trainable parameters equivalent to only about \textbf{6\%} of the target model size, avoiding full model adaptation while preserving the target model's original behavior.
\textbf{(2) \bonusbias{}} conditions on the bonus token embedding and the hidden states of mask tokens to produce a calibration bias over the draft logits.
This mechanism guides the draft distribution to be consistent with the determined bonus token, improving alignment between drafting and verification.
\textbf{(3) \flexdecoding{}} dynamically switches between parallel draft and verify at small batch sizes and sequential draft then verify at large batch sizes, while adaptively pruning the verification length based on draft confidence to eliminate redundant computation.

%% file: sec/3related_work.tex
% %dLLM
% Diffusion large language models (dLLMs) have emerged as a compelling non-autoregressive 150 paradigm for text generation. 
% enerate text by iteratively denoising partially masked sequences, predicting multiple token positions in parallel conditioned on the surrounding context.
% LLaDA LLaDA (Nie et al., 2025) was the first to scale dLLMs to billions of parameters
% SDAR dLLM-var block diffusion
% 尽管一些工作尝试进行加速，还是难以达到与自回归相匹敌的速度
% Fast-dLLM v2 (Wu et al., 2025) and SDAR (Cheng et al., 2025) adapt pre-trained autoregressive LLMs into block-diffusion variants, enabling parallel generation while preserving generation quality on specific tasks.通常伴随着质量的下降和训练成本问题
% Tidar Fast dVLM 并不能保证自回归模型性能不下降
% %投机解码
% % Eagle系列工作
% %GliDe with a CaPE: A Low-Hassle Method to Accelerate Speculative Decoding利用kv cache提高draft质量
% %dflash、dart，额外参数量巨大，draft开销高
% %先前的draft&verify存在一系列弊端，加速比极低
\textbf{Speculative Decoding}
Speculative decoding accelerates LLM inference by decomposing generation into a fast draft phase and a lossless verification phase, where a lightweight drafter proposes multiple candidate tokens and the target model verifies them in parallel using the standard speculative sampling rule \citep{chen2023accelerating, leviathan2023fast}.
Existing methods improve speculative decoding mainly by enhancing draft quality, reducing draft generation cost, or organizing candidates more efficiently for verification \citep{xia2024unlocking}.
To improve draft quality and acceptance rates, prior work has strengthened draft target alignment and draft prediction through distillation \citep{zhou2024distillspec}, head-based speculation \citep{cai2024medusa, ankner2024hydra, zhang2024koala, li2025amphista}. 
Representative methods include Medusa style approaches, which attach trainable decoding heads to the target model \citep{cai2024medusa}, and the EAGLE series, which improves drafting with target hidden states, multi-layer feature fusion, and direct token prediction \citep{li2024eagle, li2025eagle3}.
To reduce drafter cost or latency, retrieval-based methods \citep{he2024rest, gritta2025dresd, quan-etal-2025-rasd} avoid a separately trained draft model by constructing lightweight draft sources from external datastores. 
Parallel drafting methods further reduce latency by replacing sequential autoregressive drafting with parallel proposal generation \citep{christopher2025speculative,li2025diffuspec, chen2026dflash, liu2026dart}. 
On the verification side, tree-structured speculation methods organize candidate continuations into token trees for parallel verification \citep{miao2024specinfer, li2024eagle2}.
Despite these advances, most methods still follow the conventional draft then verify schedule, where the next draft cannot start until verification finishes, causing mutual waiting and limiting end-to-end speedup \citep{liu2025pearl}. 
Recent methods attempt to alleviate this bottleneck by preparing future candidates before verification is fully resolved. BiTA \citep{lin2025bita} performs draft generation and verification in parallel within a single enhanced model, and MTP-based methods \citep{samragh2025your} speculate over multiple future positions.
However, preparing future drafts before verification is resolved introduces acceptance length and bonus token uncertainty. The drafter does not know how many draft tokens will be accepted, nor which continuation token will be determined by the verifier. 

\textbf{Diffusion Language Models}
%引出dLLM，讲解dLLM
%提出block diffusion(AR2dLLM转化范式)
%尽管有一些各种各样优化，但是加速效果一般(dLLM-Cache、Slowfast sampling、 MaskKV)
%Tidar
Diffusion large language models (dLLMs) provide an alternative to autoregressive generation by denoising masked token sequences and predicting multiple positions in parallel. 
LLaDA \citep{nie2025llada} demonstrates that this paradigm can scale to billion parameters language models and achieve competitive performance with strong autoregressive baselines. 
However, fully parallel dLLMs often suffer from fixed length generation \citep{wu2026dreamon}, limited KV cache support \citep{wu2025fastdllm, liu2025dllmcache, huang2025maskkv}, and many denoising steps \citep{wu2025fastdllm, wei2025slowfast}, which restrict their practical inference speed. 
Block diffusion models \citep{arriola2025block} alleviate these limitations by generating text block by block.
Following this direction, recent work adapts autoregressive LLMs into block diffusion language models to improve parallelism, but typically requires substantial continued training and may introduce quality trade-offs \citep{wu2025fastdllmv2, cheng2025sdar}.
More recently, speculative decoding has also been incorporated into block diffusion to improve its performance in highly parallel generation settings.
For example, hybrid diffusion and autoregressive methods such as TiDAR \citep{liu2025tidar} combine diffusion style drafting with autoregressive sampling, but they lack the same target distribution guarantee as standard speculative decoding.

%% file: sec/preliminaries.tex
% Speculative decoding accelerates autoregressive generation by decoupling each decoding step into a \emph{draft phase} and a \emph{verification phase}. 
% Let $q$ denote the drafter distribution and $p_\theta$ denote the target model distribution. 
% In the draft phase, a lightweight drafter proposes $k$ candidate tokens,
% $\hat{\mathbf{y}}=(\hat{y}_{n+1},\ldots,\hat{y}_{n+k})$, to approximate the target model distribution.
% In the verification phase, the target model evaluates these candidates in a single forward pass and applies \emph{speculative sampling}~\citep{chen2023accelerating}, accepting each draft token according to the ratio between target and drafter distributions, and sampling a new token from the residual target distribution at the first rejected position.
% We refer to this newly sampled token as the \emph{bonus token}.
% The number of tokens committed in one decoding step is denoted as the \emph{accepted length} $L^{\mathrm{acc}}$, including both the accepted draft tokens and the bonus token. 
% Equivalently, if $r^{\mathrm{acc}}$ draft tokens are accepted before the bonus token is produced, then $L^{\mathrm{acc}}=r^{\mathrm{acc}}+1$. 

Speculative decoding accelerates autoregressive generation by decomposing each decoding step into drafting and verification.
Let $q$ and $p_\theta$ denote the drafter and target model distributions.
The drafter first proposes $k$ candidate tokens,
$\hat{\mathbf{y}}=(\hat{y}_{n+1},\ldots,\hat{y}_{n+k})$, to approximate $p_\theta$.
The target model then verifies them in a single forward pass using \emph{speculative sampling}~\citep{chen2023accelerating}, accepting draft tokens according to the ratio between $p_\theta$ and $q$ and sampling from the residual target distribution at the first rejected position.
We call this newly sampled token the \emph{bonus token}.
If $r^{\mathrm{acc}}$ draft tokens are accepted before the bonus token is produced, the committed length is $L^{\mathrm{acc}}=r^{\mathrm{acc}}+1$.

\noindent \textbf{Sequential speculative decoding.}
In the standard draft then verify schedule, a new draft begins only after verification completes.
Given the accepted draft token count $r^{\mathrm{acc}}$ and the bonus token $b$, the drafter conditions on the complete verified context to generate the next candidate block:
\begin{equation}
\hat{\mathbf{y}}^{\mathrm{next}}
\sim q\left(\cdot \mid \mathbf{x}_{1:n}, \hat{\mathbf{y}}_{1:r^{\mathrm{acc}}}, b\right).
\end{equation}
\noindent \textbf{Parallel speculative decoding.}
In contrast, parallel speculative decoding prepares future candidates before the current verification result is fully resolved.
Since the accepted length $L^{\mathrm{acc}}$ and the bonus token $b$ are unknown at draft time, the drafter cannot condition on the final verified context.
Instead, it must prepare candidate drafts for multiple possible numbers of accepted draft tokens.
Let $\mathcal{R}=\{0,1,\ldots,k\}$ denote the set of possible accepted draft token counts.
An idealized formulation is to construct a candidate group indexed by $r\in\mathcal{R}$:
\begin{equation}
    \mathcal{G}
    =
    \left\{
    \hat{\mathbf{y}}_{r}^{\mathrm{next}}
    \sim
    q\!\left(
    \cdot \mid
    \mathbf{x}_{1:n},
    \hat{\mathbf{y}}_{1:r}
    \right)
    \right\}_{r\in\mathcal{R}} .
    \label{eq:parallel}
\end{equation}
The case $r=0$ corresponds to immediate rejection, where no draft token is accepted and $\hat{\mathbf{y}}_{1:0}$ is defined as an empty prefix.
Each branch $\hat{\mathbf{y}}_{r}^{\mathrm{next}}$ is generated under the hypothesis that the first $r$ draft tokens are accepted and form the verified prefix for the next draft.
Once verification completes, $r^{\mathrm{acc}}$ is determined and the corresponding branch $\hat{\mathbf{y}}_{r^{\mathrm{acc}}}^{\mathrm{next}}$ is selected from $\mathcal{G}$.

\begin{figure*}[t]
  \centering
  \includegraphics[width=1.0\linewidth]{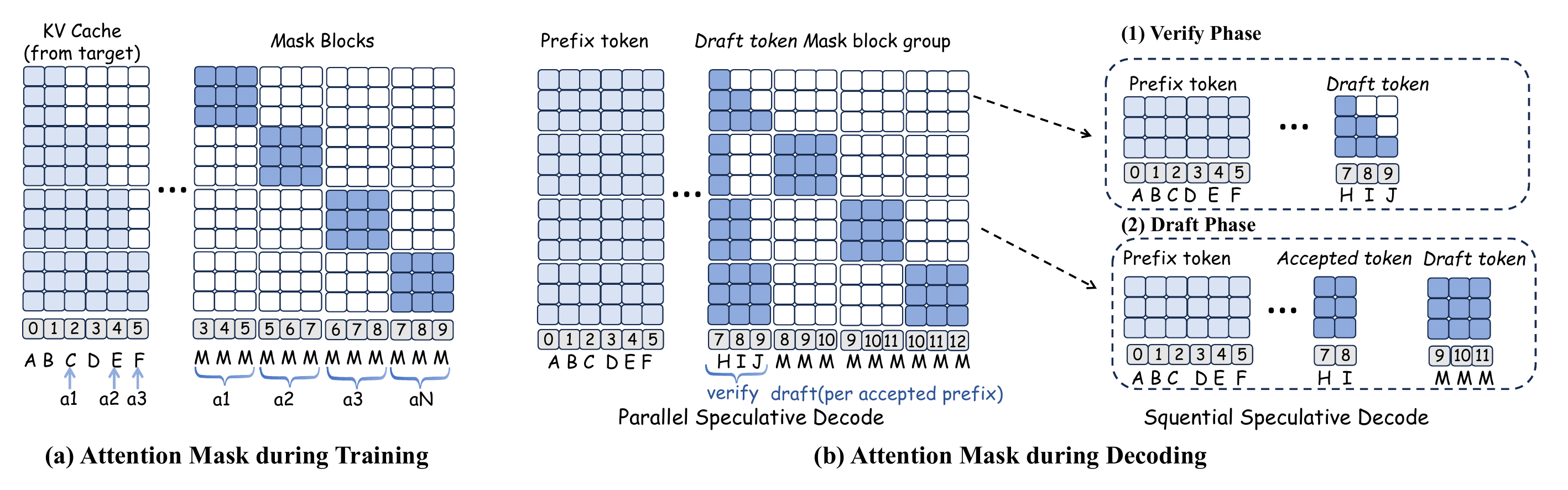}
  \caption{\textbf{Attention masks in FlexDraft.}
  \textbf{(a) Training.} The target performs causal forward to build the clean prefix KV cache. Mask tokens attend bidirectionally within each block and to the prefix, isolated from other blocks.
  \textbf{(b) Decoding.} Our method supports both parallel and sequential speculative decoding. In parallel mode, the latest draft is verified while candidates for all possible accepted lengths are prepared in the same forward pass. In sequential mode, drafting and verification are executed sequentially by selecting the verification row and its corresponding draft row from the parallel attention mask.}
  \label{fig:attn_mask}
  \vspace{-4mm}
\end{figure*}

%% file: sec/4method.tex
\mymethod{} is a lossless speculative decoding framework that combines lightweight block diffusion drafting, bonus-aware calibration, and batch-adaptive execution.
It targets both the mutual waiting and memory access overhead of sequential speculative decoding at small batch sizes and the redundant compute overhead of parallel speculative decoding at large batch sizes.
\attntuning{} enables block diffusion drafting with only a small number of additional attention projector parameters while preserving the target model's autoregressive distribution.
\bonusbias{} reduces the mismatch between bonus-free drafting and bonus-constrained verification, and \flexdecoding{} dynamically switches decoding modes and prunes verification length based on draft confidence to ensure robust acceleration across batch sizes.
Together, these designs preserve the target distribution, improve draft quality, and prevent speedup collapse under large batch serving.
\begin{figure*}[t]
  \centering
  \includegraphics[width=1.0\linewidth]{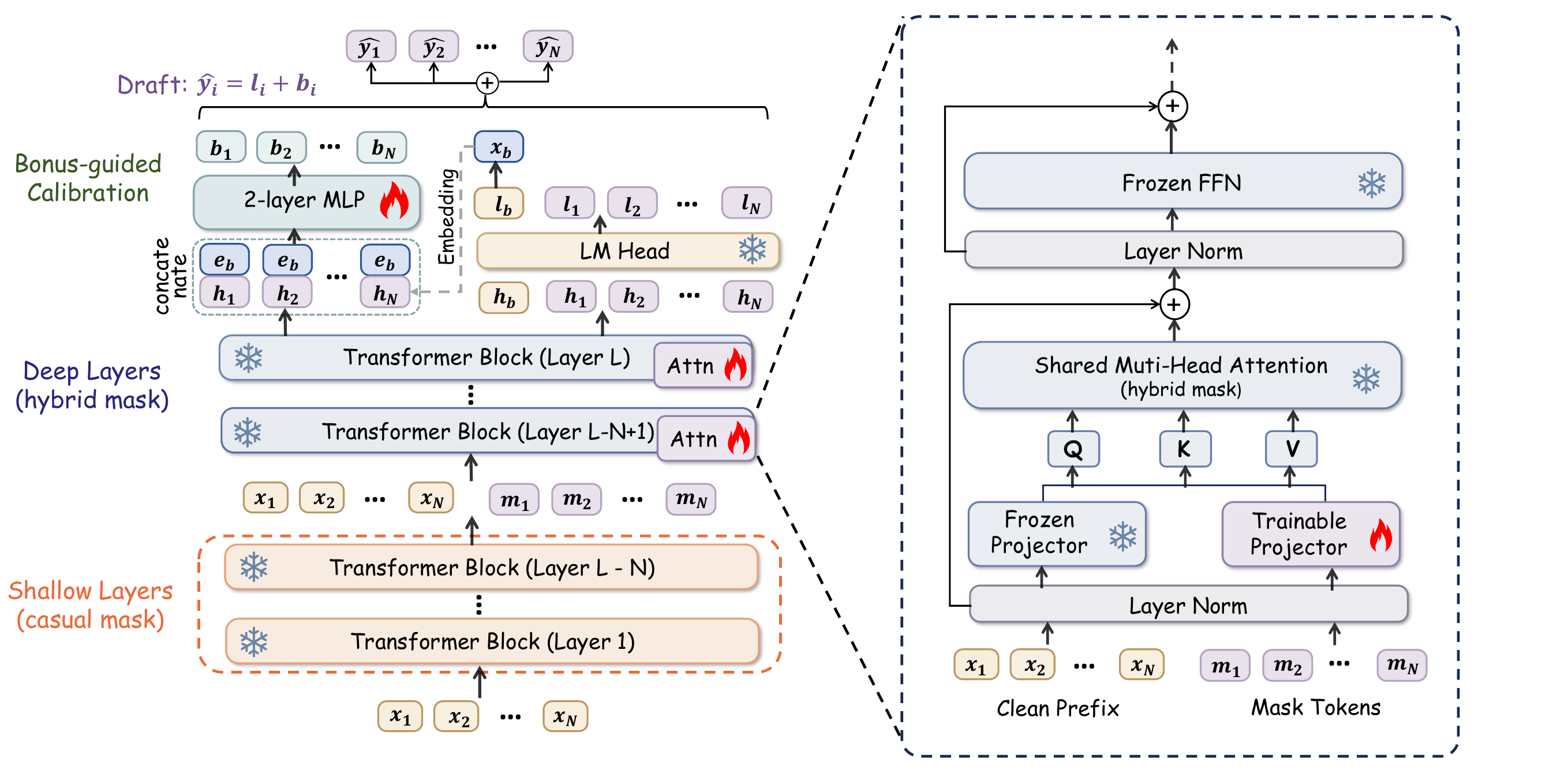}
  \caption{\textbf{Pipeline of FlexDraft.} 
  \vspace{-0.7mm}
  Shallow layers process the clean prefix identically to a standard autoregressive forward pass. In the deep layers, mask tokens are appended to the prefix and routed through trainable attention projectors, enabling parallel draft prediction. Bonus-guided Calibration injects the verified bonus token embedding into a lightweight MLP to adjust draft logits, which improves draft quality.}
  \vspace{-4mm}
  \label{fig:pipe_figure}
\end{figure*}
\subsection{Attention Tuning}
%轻量转化，额外参数量少\FFN知识
%无损
Our goal is to enable block diffusion drafting that produces $k$ draft tokens in a single forward pass while preserving the target model's autoregressive behavior, ensuring that our method remains lossless.
Prior work on multi token prediction suggests that pretrained autoregressive models already encode information about future tokens in their intermediate representations, implying that full model continual pretraining may be unnecessary \citep{samragh2025your}.
Motivated by this observation, we hypothesize that adapting only the attention projectors is sufficient to elicit parallel masked prediction, while keeping the FFN layers frozen allows the drafter to directly inherit the pretrained knowledge of the target model.
In our design, mask tokens attend bidirectionally within each block and to the clean prefix, while remaining isolated from other blocks and invisible to clean prefix tokens, thereby preventing draft positions from interfering with prefix computation.

% \noindent \textbf{Dual attention projectors.}
% Let $\mathbf{x}_{1:n}$ denote the clean prefix, comprising the prompt and all previously verified response tokens. We denote the original attention projectors of layer $l$ collectively as $\mathbf{W}^l = (W_Q^l, W_K^l, W_V^l, W_O^l)$, and introduce a dedicated set of mask projectors $\tilde{\mathbf{W}}^l$ for the final $N$ layers. For the first $L{-}N$ layers, the model processes $\mathbf{x}_{1:n}$ identically to a standard autoregressive forward pass, with no mask tokens involved, producing clean KV representations that can be cached and reused. Starting from layer $L{-}N{+}1$, $k$ identical mask embeddings $\mathbf{e}_m \in \mathbb{R}^d$ are appended to the hidden states of $\mathbf{x}_{1:n}$ to serve as draft query positions. The projector applied at position $i$ in layer $l$ is determined by whether the token is a clean prefix token or a mask token:
% \begin{equation}
%     \mathbf{W}^l_i =
%     \begin{cases}
%         \tilde{\mathbf{W}}^l & \text{if } i > n \;\text{ and }\; l \geq L{-}N{+}1 \\
%         \mathbf{W}^l        & \text{otherwise,}
%     \end{cases}
%     \label{eq:dual_proj}
% \end{equation}
% so that clean prefix tokens always use the frozen projectors $\mathbf{W}^l$ regardless of layer depth, while mask positions in the final $N$ layers are routed through the trainable $\tilde{\mathbf{W}}^l$. All parameters outside $\{\tilde{\mathbf{W}}^l\}_{l=L-N+1}^{L}$ remain frozen throughout training.

\noindent \textbf{Dual attention projectors.}
Let $\mathbf{x}_{1:n}$ denote the clean prefix, comprising the prompt and all previously verified response tokens.
As illustrated in Figure~\ref{fig:pipe_figure}, \mymethod{} processes the clean prefix identically to the original autoregressive model in the first $L{-}N$ layers, with no mask tokens involved.
Starting from layer $L{-}N{+}1$,  mask tokens $\mathbf{e}_m \in \mathbb{R}^d$ are appended to the hidden states of $\mathbf{x}_{1:n}$ as draft query positions.
To separate drafting from standard autoregressive computation, we keep the original attention projectors $\mathbf{W}^l=(W_Q^l,W_K^l,W_V^l,W_O^l)$ frozen for all clean prefix tokens, and introduce trainable mask specific projectors $\tilde{\mathbf{W}}^l$ only for mask tokens in the final $N$ layers.
All other parameters, including the FFN layers and the original autoregressive projectors, remain frozen throughout training.

\noindent \textbf{Training.}
To efficiently train multiple draft blocks in a single forward pass, we adopt a packed draft training strategy. 
We first process the clean sequence with the original target model under standard causal attention, obtaining reusable KV states for all clean positions. 
Each randomly sampled anchor position $x_n$ defines the start of a draft block, and the subsequent $k$ tokens $\{x_{n+1}, \ldots, x_{n+k}\}$ serve as prediction targets.
As illustrated in Figure~\ref{fig:attn_mask}, packed draft training is implemented by controlling visibility with attention masks and position IDs, rather than physically appending mask embeddings after each anchor. 
For each anchor $x_n$, its $k$ mask tokens are assigned position IDs $n{+}1$ through $n{+}k$, attend to the cached clean prefix up to $x_n$, and interact bidirectionally within the same block. 
Mask tokens from different draft blocks are mutually invisible, allowing multiple draft blocks to be packed into one forward pass without information leakage.
These mask tokens are processed by the trainable mask specific attention projectors together with the frozen FFN layers inherited from the target model. 
This allows the drafter to reuse the target model's pretrained knowledge while adapting only a small set of parameters. 
The resulting training process faithfully simulates block diffusion drafting at inference time and enables all $k$ draft tokens to be predicted in parallel by the drafter $q_\phi$.
Following DFlash, we apply exponentially decaying weights to the block level cross entropy, emphasizing earlier draft positions because an early mismatch invalidates all subsequent draft tokens during verification. 
For a draft block starting at anchor $x_n$, the training objective is:
\begin{equation}
    \mathcal{L}_{\mathrm{draft}}
    =
    -\frac{1}{\sum_{i=1}^{k} w_i}
    \sum_{i=1}^{k}
    w_i
    \log q_\phi\!\left(
        x_{n+i} \mid \mathbf{x}_{1:n}
    \right),
    \quad
    w_i = \exp\!\left(-\lambda(i-1)\right),
    \label{eq:draft_loss}
\end{equation}
where $\lambda$ is a decay hyperparameter, set to $7$ unless otherwise specified. 
The final training loss is averaged over all sampled anchors in the packed batch.

\noindent \textbf{Inference.}
At each decoding step, \mymethod{} executes drafting and verification concurrently. 
The previous draft $\hat{\mathbf{y}}^{(t)}$ is fed into the model, and the first $L-N$ layers process the clean prefix $\mathbf{x}_{1:n}$ autoregressively using the frozen projectors $\mathbf{W}^l$, identical to a standard forward pass, producing a KV cache that can be reused across all decoding steps.
Starting from layer $L-N+1$, $k$ mask embeddings $\mathbf{e}_m^k$ are injected and routed through the trainable mask specific projectors $\tilde{\mathbf{W}}^l$, while clean prefix tokens continue through the frozen projectors.
Using the hybrid attention mask, mask tokens attend bidirectionally within their block and to the clean prefix, and multiple candidate drafts are generated in parallel for all possible hypothetical acceptance lengths $\ell \in \mathcal{L}$.
Once the actual accepted length $L^{\mathrm{acc}}$ is determined, the corresponding candidate $\hat{\mathbf{y}}^{\mathrm{next}}_{L^{\mathrm{acc}}}$ is selected for the next step.

\subsection{\bonusbias{}}
In the parallel draft and verify paradigm, drafting proceeds without the bonus token, while verification is constrained by the resolved bonus token, which causes a draft verification mismatch.
We introduce \bonusbias{}, a lightweight module applied after the LM head, to bridge this mismatch by calibrating the draft logits with the bonus token.

Let $h_i \in \mathbb{R}^d$ denote the hidden state of mask position $i$ at the final draft layer, and $e_b \in \mathbb{R}^d$ the embedding of the determined bonus token $b^{(t)}$. The calibration bias is jointly determined by both the bonus token and the draft content at each position. 
Thus, \bonusbias{} concatenates $e_b$ with $h_i$ and passes the result through a lightweight two layer MLP to produce an additive bias over the vocabulary logits:
\begin{equation}
    \tilde{\boldsymbol{\ell}}_i
    =
    \boldsymbol{\ell}_i
    +
    \mathrm{MLP}\!\left(
    \mathrm{Concat}(\mathbf{e}_b, \mathbf{h}_i)
    \right),
    \label{eq:bonusbias}
\end{equation}
where $\ell_i \in \mathbb{R}^{|\mathcal{V}|}$ are the original draft logits at position $i$ produced by the language modeling head. 
The calibrated logits $\tilde{\ell}_i$ are used to produce the final draft tokens, steering the draft distribution toward continuations consistent with $b^{(t)}$. During training, we use the ground truth token embedding as a proxy for $e_b$, and train the MLP jointly with \attntuning{} under an additional cross entropy loss, adding negligible parameter overhead.

\subsection{\flexdecoding{}}
\flexdecoding{} adapts the execution mode to the dominant bottleneck at different batch sizes, enabling a single trained drafter to support both inference modes without retraining.
At small batch sizes, decoding is largely memory bound, so parallel draft and verify is preferable for avoiding mutual waiting bubbles and intermediate state exchange overhead.
At large batch sizes, the target forward pass becomes compute bound, and the redundant branches of parallel speculative decoding make sequential draft then verify more efficient.
Unlike prior parallel methods that rely on full model drafting, our drafter is localized to the final few layers through mask specific attention projectors and is therefore detachable.
It can be inserted into the target forward pass for parallel execution or decoupled as a lightweight sequential draft step that reuses the clean prefix KV cache from target verification.
Accordingly, \flexdecoding{} uses a batch size threshold to switch between the two modes, adopting \emph{Selective Verification} at small batch sizes to prune low confidence candidate lengths and switching to \emph{Decoupled Execution} at large batch sizes to generate only the branch conditioned on the resolved accepted prefix.

\noindent \textbf{Selective verification (small batch).}
In the parallel mode, \mymethod{} prepares candidate drafts for multiple possible accepted lengths.
Although the accepted length is unknown before verification, draft confidence provides a useful proxy for estimating which candidate lengths are more likely to be selected.
Intuitively, higher confidence draft tokens are more likely to be accepted, and under a simple independence approximation across positions, the probability that the accepted length reaches position $i$ can be estimated by the cumulative confidence.
Let $p_i = q_\phi(\hat{y}_i \mid \mathbf{x}_{1:n})$ denote the draft probability of the $i$-th drafted token.
We approximate this likelihood as
\begin{equation}
    P(L^{\mathrm{acc}} \geq i) \approx \prod_{r=1}^{i} p_r .
\label{eq:selective_verify}
\end{equation}
In this way, \mymethod{} converts complete accepted length uncertainty into a coarse confidence-based estimate.
Once the cumulative confidence falls below a threshold, we prune the remaining low probability accepted length candidates and verify only a subset of high confidence prefixes.
This reduces the number of candidate prefixes verified by the target model without changing the draft block length, so each retained branch still preserves the same $k$ token drafting capacity.

% \noindent \textbf{Decoupled execution (large batch).}
% As batch size increases, the verification cost grows proportionally and the system enters the compute-bound regime. Maintaining the full candidate group $\mathcal{G}$ over $\mathcal{L}$ in this setting introduces redundant branch computation proportional to $|\mathcal{L}|$, which degrades throughput and eventually causes acceleration gains to collapse. \flexdecoding{} addresses this by switching to a sequential draft-then-verify strategy, where drafting is conditioned on the resolved accepted prefix $\mathbf{x}_{1:n+L^{\mathrm{acc}}}$ and a single block diffusion step generates the next $k$ draft tokens, collapsing the candidate group to a singleton and eliminating all redundant computation. Since the clean prefix computation under Eq.~\eqref{eq:attn_mask} is identical to a standard autoregressive forward pass, the KV cache produced during target verification can be directly reused for drafting without recomputation, keeping the sequential draft step lightweight. This design effectively prevents the throughput collapse observed in parallel speculative decoding at large batch sizes.

\noindent \textbf{Decoupled execution (large batch).}
At large batch sizes, the redundant branches of parallel speculative decoding dominate the cost.
Thus, \flexdecoding{} switches to decoupled execution, as shown in the sequential speculative decoding mode of Figure~\ref{fig:attn_mask}.
Once verification resolves the accepted length $L^{\mathrm{acc}}$, the drafter generates only one block conditioned on the verified prefix.
Because the clean prefix computation is identical to the target autoregressive computation, the KV cache produced during target verification can be reused for drafting.
Thus, only one branch is computed, avoiding redundant accepted length branches and preventing large batch speedup collapse.

%% file: sec/5experiment.tex
\subsection{Experimental Setup.}

We evaluate FlexDraft on Qwen3 series target models across code generation, mathematical reasoning, and general language understanding benchmarks, including HumanEval, MBPP, GSM8K, MATH, and MT-Bench. All experiments are conducted on NVIDIA A100 GPUs. We compare against representative speculative decoding baselines, including DFlash~\cite{chen2026dflash}, EAGLE-3~\cite{li2025eagle3}, DART~\cite{liu2026dart}, BiTA~\cite{lin2025bita} and Apple MTP~\cite{samragh2025your}.
We do not include TiDAR as a main baseline because it is a hybrid diffusion autoregressive method that does not generally provide the same lossless target distribution guarantee as standard speculative decoding, and its implementation has not been publicly released.
Unless otherwise specified, all experiments are conducted with batch size 1, and all trainable methods are trained on the same 300K samples from \texttt{mlabonne/open-perfectblend}\footnote{\url{https://huggingface.co/datasets/mlabonne/open-perfectblend}}.

% \noindent \textbf{Implementation.}
% We set the number of draft model layers to 10 and the block size to 16. The draft models are trained on 300K samples drawn from PerfectBlender, and optimized for 6 epochs using the AdamW optimizer with a learning rate of $6 \times 10^{-5}$, a gradient clipping threshold of 1.0, and a cosine learning rate schedule with a warmup ratio of 0.04. For each input sequence, 128 anchor positions are randomly sampled as training targets.

\subsection{Main Results}
\input{table/main_table}

\noindent \textbf{Comparison under the same data.}
For a fair comparison, all trainable methods are trained on the same data and evaluated under identical settings.
We first compare with parallel speculative decoding methods, including BiTA and Apple MTP.
Their performance is constrained by the inherent limitations of the parallel speculative decoding paradigm, as well as limited trainable capacity and low acceptance rates.
By explicitly mitigating bonus token uncertainty and reducing redundant computation, \mymethod{} substantially outperforms existing parallel speculative decoding methods.

\mymethod{} also consistently surpasses strong speculative decoding baselines, including EAGLE-3 and DFlash, across model scales and benchmarks.
The key advantage is that \mymethod{} reuses the target model's frozen FFN parameters and only introduces lightweight mask specific attention projectors for draft tokens.
As a result, increasing the draft depth incurs only small parameter overhead while producing longer accepted continuations.
This enables \mymethod{} to achieve higher acceptance lengths and speedups, while retaining the parallel drafting benefit that removes mutual waiting bubbles at small batch sizes.

% \noindent \textbf{Comparison under scaled training.}
% \input{table/scale_data}
% Table~\ref{tab:qwen3_8b_released} further compares \mymethod{} with released or publicly reported SOTA methods on Qwen3-8B in a scaled training setting to examine its scaling potential.
% \mymethod{} clearly outperforms DART and EAGLE-3 across most benchmarks, while achieving slightly better or comparable performance to DFlash.
% The smaller margin over DFlash compared with Table~\ref{tab:qwen3_performance_final} is mainly because DFlash does not open-source its training data, making a fully data-matched comparison infeasible.

\noindent \textbf{Comparison under scaled training.}
\input{table/scale_data}
Table~\ref{tab:qwen3_8b_released} further compares \mymethod{} with released or publicly reported SOTA methods on Qwen3-8B under scaled training, evaluating its scaling potential in a more competitive setting.
\mymethod{} consistently outperforms DART and EAGLE-3 across most benchmarks.
Compared with DFlash, \mymethod{} achieves comparable or slightly better performance, despite not having access to the training data used by DFlash.
These results demonstrate that \mymethod{} scales effectively and remains competitive with strong publicly reported methods under a non-data-matched comparison.

\begin{figure*}[t]
\centering
\begin{minipage}[t]{0.31\textwidth}
    \centering
    \includegraphics[width=\linewidth]{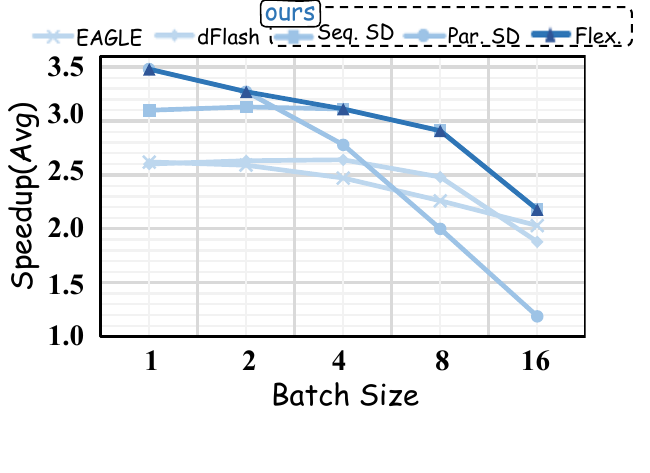}
    \vspace{-0.9cm} % <-- 调整这里的数值来控制间距
    \caption{\textbf{Speedup across batch sizes.}}
    \label{fig:batch_size_speedup}
\end{minipage}%
\hfill
\begin{minipage}[t]{0.31\textwidth}
    \centering
    \includegraphics[width=\linewidth]{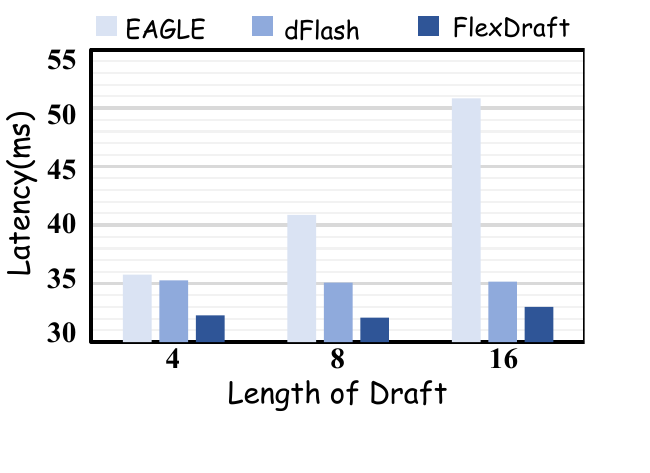}
    \vspace{-0.9cm} % <-- 调整这里的数值来控制间距
    \caption{\textbf{Execution time of a single draft and verify step.}}
    \label{fig:draft_verify_latency}
\end{minipage}%
\hfill
\begin{minipage}[t]{0.31\textwidth}
    \centering
    \includegraphics[width=\linewidth]{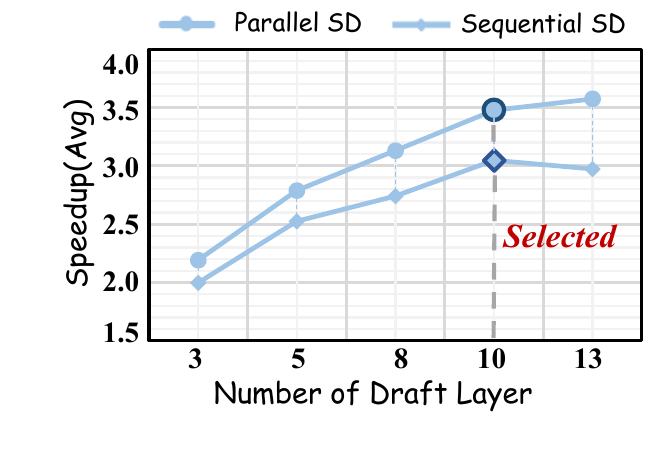}
    \vspace{-0.9cm} % <-- 调整这里的数值来控制间距
    \caption{\textbf{Effect of the number of draft layers.}}
    \label{fig:num_draft_layers}
\end{minipage}
\vspace{-0.4cm}
\end{figure*}

% \noindent \textbf{Batch size study.}
% Figure~\ref{fig:batch_size_speedup} evaluates performance under different batch sizes.
% At small batch sizes, decoding is largely memory-bound, and sequential speculative decoding suffers from mutual-waiting bubbles between drafting and verification.
% \mymethod{} mitigates this issue by executing the two stages in parallel.
% As batch size increases, the target forward pass becomes more compute-bound and dominates end-to-end latency, reducing the relative impact of mutual waiting.
% The main bottleneck then shifts to the redundant target-forward computation introduced by parallel speculative decoding, whose effective token count grows with the number of possible acceptance prefixes.
% To address this, \mymethod{} falls back to sequential draft-then-verify execution at large batch sizes, where the accepted prefix is already resolved and only a single draft branch is generated.
% This allows \mymethod{} to exploit parallelism in small-batch regimes while avoiding excessive forward overhead in large-batch regimes.
% Seq. SD和Par. SD是我们方法训练之后的两种不同推理方案(一种训练两种推理方案)可根据batch size进行选择，当batch size较小时Par.SD加速比更高，当batch size较大的时Seq. SD，我们的方法FlexDraft在batch以2为阈值，当大于2的时候采用Seq的方式可以避免小batch的气泡和大batch的overhead。我们的方法由比dflash和Eagle3好

\noindent \textbf{Batch size study.}
Figure~\ref{fig:batch_size_speedup} evaluates the performance under different batch sizes.
Seq. SD and Par. SD are two inference modes derived from the same trained \mymethod{} model, rather than separately trained variants.
At small batch sizes, parallel speculative decoding achieves higher speedup because drafting and verification can be overlapped, effectively reducing the mutual waiting bubbles between the two stages.
However, as the batch size increases, the redundant computation introduced by parallel speculative decoding becomes more pronounced, making sequential speculative decoding more favorable by drafting only along the resolved accepted branch.
Based on this observation, we use a batch size of 2 as the threshold for switching from parallel to sequential speculative decoding in \mymethod{}.
This adaptive strategy allows \mymethod{} to benefit from parallel execution in small batch regimes while avoiding excessive overhead in large batch regimes.
As a result, \mymethod{} consistently achieves higher speedup than EAGLE-3 and dFlash across batch sizes, demonstrating both the effectiveness of our training design and the necessity of batch-aware inference selection.

\subsection{Ablation}
\begin{figure*}[t]
\centering
\begin{minipage}[t]{0.26\textwidth}
    \centering
    \includegraphics[width=\linewidth]{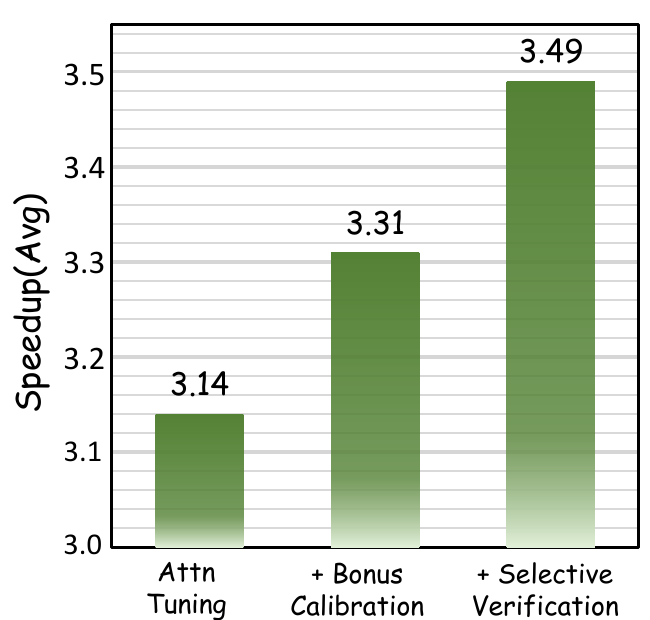}
    \caption{\textbf{Ablation analysis of speedup.}}
    \label{fig:component_speedup}
\end{minipage}%
\hfill
\begin{minipage}[t]{0.72\textwidth}
    \centering
    \includegraphics[width=\linewidth]{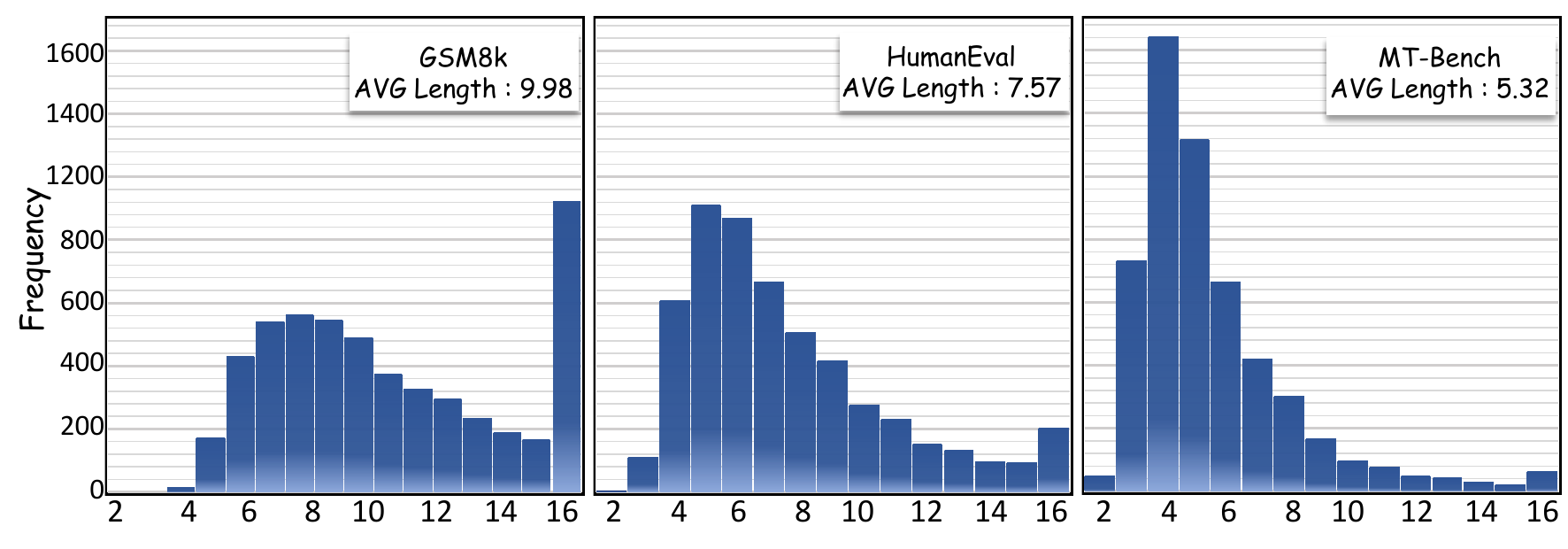}
    \caption{\textbf{Distribution of selective verification length.}
    The selected verification length varies across tasks.}
    \label{fig:verify_length}
\end{minipage}
\vspace{-5mm}
\end{figure*}
%研究一下forward增加的额外开销
\noindent \textbf{Latency of a single draft and verify step.}
Figure~\ref{fig:draft_verify_latency} compares the latency of one draft and verify step.
EAGLE and dFlash first generate draft tokens and then verify them with the target model, so their latency accumulates across the two stages.
In contrast, \mymethod{} avoids a separate autoregressive drafting pass.
It generates a block of draft tokens in one pass while target verification is performed, thereby avoiding the sequential draft then verify pipeline and reducing per step latency.

\noindent \textbf{Effect of the number layer of draft model.}
Figure ~\ref{fig:num_draft_layers} studies the number of draft layers.
A larger draft depth generally improves acceptance length by strengthening draft target alignment.
While 13 layers yields slightly higher speedup in the parallel setting, it increases trainable parameters and draft computation, which becomes less favorable under decoupled execution.
We therefore use 10 layers by default, as it provides a better trade-off between speedup and decoding overhead.
%尽管10层比dflash大，额外参数量少、acc提升大于draft overhead

\noindent \textbf{Effectiveness of components.}
Figure~\ref{fig:component_speedup} shows the contribution of each component.
Starting from Attn Tuning, adding Bonus-guided Calibration improves the average speedup by better aligning drafts with the bonus-constrained verification.
Selective Verification further improves speedup by pruning low confidence verification positions and reducing redundant target forward computation.
These results show that the two components are effective.

%% file: table/main_table.tex
\begin{table}[t]
\centering
\caption{Decoding performance on Qwen3 models. $\tau$ denotes average acceptance length. The highest speedup for each benchmark is highlighted in \textbf{bold}.}
\label{tab:qwen3_performance_final}
\renewcommand{\arraystretch}{1.2}
\resizebox{\textwidth}{!}{%
\begin{tabular}{ll cc cc cc cc cc}
\toprule
\multirow{2}{*}{Model} & \multirow{2}{*}{Method}
  & \multicolumn{2}{c}{GSM8K}
  & \multicolumn{2}{c}{MATH}
  & \multicolumn{2}{c}{HumanEval}
  & \multicolumn{2}{c}{MBPP}
  & \multicolumn{2}{c}{MT-Bench} \\
\cmidrule(lr){3-4}\cmidrule(lr){5-6}\cmidrule(lr){7-8}\cmidrule(lr){9-10}\cmidrule(lr){11-12}
& & $\tau$ & Speedup & $\tau$ & Speedup & $\tau$ & Speedup & $\tau$ & Speedup & $\tau$ & Speedup \\
\midrule

% --- Qwen3-8B ---
\multirow{5}{*}{Qwen3-8B} 
  & BiTA      & 1.68 & 1.34$\times$ & 1.63 & 1.30$\times$ & 1.49 & 1.21$\times$ & 1.47 & 1.19$\times$ & 1.47 & 1.15$\times$ \\
  & Apple MTP & 3.30 & 2.77$\times$ & 3.14 & 2.61$\times$ & 2.49 & 2.09$\times$ & 2.44 & 2.04$\times$ & 2.11 & 1.63$\times$ \\
  & EAGLE-3   & 4.84 & 3.40$\times$ & 4.88 & 3.34$\times$ & 3.57 & 2.47$\times$ & 3.48 & 2.45$\times$ & 2.81 & 1.75$\times$ \\
  & dFlash    & 4.87 & 3.47$\times$ & 4.69 & 3.27$\times$ & 3.10 & 2.28$\times$ & 3.23 & 2.33$\times$ & 2.53 & 1.64$\times$ \\
\rowcolor{gray!15}\cellcolor{white}
  & \textbf{FlexDraft} & \textbf{6.12} & \textbf{4.57}$\times$ & \textbf{5.98} & \textbf{4.40}$\times$ & \textbf{4.15} & \textbf{3.25}$\times$ & \textbf{3.93} & \textbf{3.04}$\times$ & \textbf{3.22} & \textbf{2.13}$\times$ \\

\midrule

% --- Qwen3-4B ---
\multirow{3}{*}{Qwen3-4B} 
  & EAGLE-3   & 4.19 & 2.82$\times$ & 4.03 & 2.82$\times$ & 3.01 & 2.09$\times$ & 2.98 & 2.08$\times$ & 2.45 & 1.53$\times$ \\
  & dFlash    & 5.18 & 3.66$\times$ & 4.93 & 3.42$\times$ & 3.41 & 2.47$\times$ & 3.45 & 2.46$\times$ & 2.78 & 1.76$\times$ \\
\rowcolor{gray!15}\cellcolor{white}
  & \textbf{FlexDraft} & \textbf{5.28} & \textbf{4.04}$\times$ & \textbf{5.35} & \textbf{4.02}$\times$ & \textbf{3.73} & \textbf{2.93}$\times$ & \textbf{3.56} & \textbf{2.82}$\times$ & \textbf{3.01} & \textbf{1.99}$\times$ \\

\midrule

% --- Qwen3-1.7B ---
\multirow{3}{*}{Qwen3-1.7B} 
  & EAGLE-3   & 3.31 & 2.12$\times$ & 3.36 & 2.14$\times$ & 2.39 & 1.49$\times$ & 2.41 & 1.51$\times$ & 1.96 & 1.31$\times$ \\
  & dFlash    & 4.87 & 3.25$\times$ & 4.58 & 3.01$\times$ & 3.20 & 2.20$\times$ & 3.17 & 2.16$\times$ & 2.72 & 1.68$\times$ \\
\rowcolor{gray!15}\cellcolor{white}
  & \textbf{FlexDraft} & \textbf{4.98} & \textbf{3.67}$\times$ & \textbf{4.94} & \textbf{3.61}$\times$ & \textbf{3.41} & \textbf{2.61}$\times$ & \textbf{3.26} & \textbf{2.48}$\times$ & \textbf{3.03} & \textbf{1.99}$\times$ \\

\bottomrule
\end{tabular}%
}
\vspace{-5mm}
\end{table}

%% file: table/scale_data.tex
\begin{table}[t]
\centering
\caption{Decoding performance on Qwen3-8B. $\tau$ denotes average acceptance length. The highest speedup for each benchmark is highlighted in \textbf{bold}.}
\label{tab:qwen3_8b_released}
\renewcommand{\arraystretch}{1.2}
\resizebox{\textwidth}{!}{%
\begin{tabular}{ll cc cc cc cc cc}
\toprule
\multirow{2}{*}{Model} & \multirow{2}{*}{Method}
  & \multicolumn{2}{c}{GSM8K}
  & \multicolumn{2}{c}{MATH}
  & \multicolumn{2}{c}{HumanEval}
  & \multicolumn{2}{c}{MBPP}
  & \multicolumn{2}{c}{MT-Bench} \\
\cmidrule(lr){3-4}\cmidrule(lr){5-6}\cmidrule(lr){7-8}\cmidrule(lr){9-10}\cmidrule(lr){11-12}
& & $\tau$ & Speedup & $\tau$ & Speedup & $\tau$ & Speedup & $\tau$ & Speedup & $\tau$ & Speedup \\
\midrule

\multirow{5}{*}{Qwen3-8B} 
  & Apple MTP & 3.30 & 2.77$\times$ & 3.14 & 2.61$\times$ & 2.49 & 2.09$\times$ & 2.44 & 2.04$\times$ & 2.11 & 1.63$\times$ \\
  & DART      & 2.71 & 2.28$\times$ & 2.70 & 2.29$\times$ & 2.95 & 2.52$\times$ & 2.98 & 2.39$\times$ & 3.03 & 2.27$\times$ \\
  & EAGLE-3   & 3.80 & 2.57$\times$ & 3.61 & 2.44$\times$ & 3.74 & 2.52$\times$ & 3.31 & 2.24$\times$ & 3.04 & 1.89$\times$ \\
  & dFlash    & 6.41 & 4.55$\times$ & 7.93 & 5.50$\times$ & 6.47 & \textbf{4.68}$\times$ & 5.93 & 4.16$\times$ & 4.29 & 2.53$\times$ \\
\rowcolor{gray!15}\cellcolor{white}
  & \textbf{FlexDraft} & 7.98 & \textbf{5.88}$\times$ & 8.03 & \textbf{5.79}$\times$ & 5.55 & 4.26$\times$ & 6.16 & \textbf{4.46}$\times$ & 4.16 & \textbf{2.58}$\times$ \\

\bottomrule
\end{tabular}%
}
\vspace{-4mm}
\end{table}

%% file: sec/6discussion.tex
\noindent \textbf{Adaptive verification length across tasks.}
Figure~\ref{fig:verify_length} shows the distribution of selective verification length across tasks.
The average is correlated with the average acceptance length, where tasks with longer accepted continuations receive more verification.
This indicates that selective verification performs sample-wise dynamic verification instead of using a fixed verification length, which can exploit potential speedup while avoiding redundant target forward computation.

\begin{wraptable}{r}{0.45\textwidth} % {r}表示靠右，0.45\textwidth是表格占的宽度
    \centering
    \caption{Effect of target knowledge reuse.}
    \label{tab:ffn_reuse}
    \renewcommand{\arraystretch}{1.1}
    \small % 缩小字号以适应窄列
    \setlength{\tabcolsep}{4pt} % 缩小列间距
    \begin{tabular}{ll cc}
    \toprule
    \textbf{Benchmark} & \textbf{Setting} & $\bm{\tau}$ & \textbf{Speedup} \\
    \midrule
    \multirow{2}{*}{GSM8K}     & Full param   & 3.78 & 3.04 \\
                               & Attn Tuning  & \textbf{5.86} & \textbf{4.23} \\
    \midrule
    \multirow{2}{*}{HumanEval} & Full param   & 2.57 & 2.10 \\
                               & Attn Tuning  & \textbf{3.99} & \textbf{2.94} \\
    \midrule
    \multirow{2}{*}{MT-Bench}  & Full param   & 2.14 & 1.60 \\
                               & Attn Tuning  & \textbf{3.06} & \textbf{1.90} \\
    \bottomrule
    \end{tabular}
\end{wraptable}

\noindent \textbf{Reusing target knowledge for drafting.}
Attn Tuning only finetunes the attention projectors while reusing the target model's frozen FFN, allowing the drafter to inherit pretrained knowledge from target model.
To analyze this effect, we compare Attn Tuning with a fully finetuned drafter under the same trainable parameter budget.
The full parameter drafter relies only on its trained parameters at inference time, whereas Attn Tuning can additionally leverage the frozen FFN of the target model.
As shown in Table~\ref{tab:ffn_reuse}, this design consistently improves both acceptance length and speedup across benchmarks, demonstrating the benefit of reusing target knowledge for drafting.

%% file: sec/7conclusion.tex
We introduced FlexDraft, a lossless speculative decoding framework that combines lightweight block diffusion drafting with adaptive execution. By tuning only attention projectors, calibrating drafts with the resolved bonus token, and dynamically switching speculative decoding strategies across batch sizes, FlexDraft reduces both mutual waiting bubbles and redundant overhead. Experiments on Qwen3 models demonstrate consistent speedups over strong baselines across reasoning, coding, and chat benchmarks, while preserving the target model distribution. FlexDraft therefore offers a practical path toward efficient LLM inference under diverse serving conditions.

%% file: sec/8appendix.tex
\section{Appendix}

\subsection{Robustness to sampling temperature.}
\input{table/temperature}
We further evaluate robustness under different sampling temperatures, i.e., $T=0.3, 0.6, 1.0$.
Table~\ref{tab:temperature} shows that \mymethod{} remains competitive across all settings and achieves the best speedup in most cases, indicating stable performance of our method.

%% file: table/temperature.tex
\begin{table}[h]
\centering
\caption{Decoding performance on Qwen3-8B under different temperatures ($T=0.3, 0.6, 1.0$). $\tau$ denotes average acceptance length. The highest speedup for each benchmark is highlighted in \textbf{bold}.}
\label{tab:temperature}
\renewcommand{\arraystretch}{1.2}
\resizebox{\textwidth}{!}{%
\begin{tabular}{cl cc cc cc cc cc}
\toprule
\multirow{2}{*}{Temp} & \multirow{2}{*}{Method}
  & \multicolumn{2}{c}{GSM8K}
  & \multicolumn{2}{c}{MATH}
  & \multicolumn{2}{c}{HumanEval}
  & \multicolumn{2}{c}{MBPP}
  & \multicolumn{2}{c}{MT-Bench} \\
\cmidrule(lr){3-4}\cmidrule(lr){5-6}\cmidrule(lr){7-8}\cmidrule(lr){9-10}\cmidrule(lr){11-12}
& & $\tau$ & Speedup & $\tau$ & Speedup & $\tau$ & Speedup & $\tau$ & Speedup & $\tau$ & Speedup \\
\midrule

% --- T = 0.3 ---
\multirow{3}{*}{$T=0.3$} 
  & dFlash    & 4.83 & 3.44$\times$ & 4.64 & 3.24$\times$ & 3.09 & 2.26$\times$ & 3.21 & 2.32$\times$ & 2.51 & 1.62$\times$ \\
  & EAGLE-3   & 4.67 & 3.27$\times$ & 4.61 & 3.21$\times$ & 3.58 & 2.42$\times$ & 3.53 & 2.41$\times$ & 2.44 & 1.74$\times$ \\
\rowcolor{gray!15}\cellcolor{white}
  & \textbf{\mymethod{}} & 5.61 & \textbf{4.20}$\times$ & 5.45 & \textbf{3.99}$\times$ & 3.85 & \textbf{2.99}$\times$ & 3.58 & \textbf{2.75}$\times$ & 2.92 & \textbf{1.92}$\times$ \\

\midrule

% --- T = 0.6 ---
\multirow{3}{*}{$T=0.6$} 
  & dFlash    & 4.63 & 3.34$\times$ & 4.48 & 3.12$\times$ & 3.01 & 2.21$\times$ & 3.14 & 2.28$\times$ & 2.50 & 1.60$\times$ \\
  & EAGLE-3   & 4.60 & 3.21$\times$ & 4.36 & 3.11$\times$ & 3.44 & 2.36$\times$ & 3.47 & 2.36$\times$ & 2.37 & 1.70$\times$ \\
\rowcolor{gray!15}\cellcolor{white}
  & \textbf{\mymethod{}} & 5.11 & \textbf{3.83}$\times$ & 4.91 & \textbf{3.56}$\times$ & 3.46 & \textbf{2.70}$\times$ & 3.34 & \textbf{2.60}$\times$ & 2.66 & \textbf{1.81}$\times$ \\

\midrule

% --- T = 1.0 ---
\multirow{3}{*}{$T=1.0$} 
  & dFlash    & 4.46 & 3.18$\times$ & 4.23 & 2.93$\times$ & 2.92 & 2.14$\times$ & 3.06 & 2.20$\times$ & 2.36 & 1.56$\times$ \\
  & EAGLE-3   & 4.41 & 3.12$\times$ & 3.96 & 2.88$\times$ & 3.24 & 2.26$\times$ & 3.21 & 2.27$\times$ & 2.27 & \textbf{1.67}$\times$ \\
\rowcolor{gray!15}\cellcolor{white}
  & \textbf{\mymethod{}} & 4.61 & \textbf{3.43}$\times$ & 4.61 & \textbf{3.44}$\times$ & 4.33 & \textbf{3.09}$\times$ & 2.99 & \textbf{2.29}$\times$ & 2.41 & 1.65$\times$ \\

\bottomrule
\end{tabular}%
}
\end{table}